\documentclass{article}
\usepackage{spconf}
\usepackage{graphicx}
\usepackage{amssymb,amsmath,bm,bbm}
\usepackage{textcomp}
\usepackage[inline]{enumitem}
\usepackage{graphics}
\usepackage{multirow}
\usepackage{hyperref}

\title{Unsupervised Domain Adaptation for Robust Speech Recognition\\via Variational Autoencoder-Based Data Augmentation}
\name{Wei-Ning Hsu\sthanks{This research was supported by a TUSA Fellowship and by PingAn.}, Yu Zhang, James Glass}
\address{Massachusetts Institute of Technology 
         Computer Science and Artificial Intelligence Laboratory \\
         Cambridge, MA 02139, USA\\
         \texttt{\{wnhsu,yzhang87,glass\}@csail.mit.edu}}
\begin{document}
\ninept
\maketitle
\begin{abstract}
Domain mismatch between training and testing can lead to significant degradation in performance in many machine learning scenarios. 
Unfortunately, this is not a rare situation for automatic speech recognition deployments in real-world applications. 
Research on robust speech recognition can be regarded as trying to overcome this domain mismatch issue.
In this paper, we address the unsupervised domain adaptation problem for robust speech recognition, where both source and target domain speech are available, but word transcripts are only available for the source domain speech.
We present novel augmentation-based methods that transform speech in a way that does not change the transcripts.
Specifically, we first train a variational autoencoder on both source and target domain data (without supervision) to learn a latent representation of speech. 
We then transform nuisance attributes of speech that are irrelevant to recognition by modifying the latent representations, in order to augment labeled training data with additional data whose distribution is more similar to the target domain.
The proposed method is evaluated on the CHiME-4 dataset and reduces the absolute word error rate (WER) by as much as 35\% compared to the non-adapted baseline.
\end{abstract}
\section{Introduction}
\label{sec:intro}
Recent advances in neural network-based acoustic models~\cite{sak2014long,sainath2015deep,peddinti2015time,hsu2016prioritized,sak2015acoustic} have greatly improved the performance of automatic speech recognition (ASR) systems, enabling more applications to adopt speech-based human-machine interaction. 
With the increasing use of ASR systems in everyday life, ASR robustness under adverse conditions becomes more essential than ever.
Some robust ASR research focuses on enhancing speech, by applying beam-forming techniques~\cite{anguera2007acoustic,erdogan2016multi}, estimating noise masks~\cite{narayanan2013ideal,isik2016single}, or training denoising models~\cite{maas2012recurrent,feng2014speech}, etc.
Other research extracts robust acoustic features~\cite{kingsbury1998robust,stern2012features,vinyals2011comparing,sainath2012auto} that are intended to be invariant for ASR even in adverse environments.
Another line of research investigates modeling techniques, including, but not limited to, model adaptation~\cite{gales1998maximum,yu2013kl}, and training models on data in adverse conditions~\cite{seltzer2013investigation,sun2017unsupervised}.
Over the past decade, neural network-based acoustic models have come to dominate the ASR field.  To utilize the full capacity of neural network-based acoustic models, it is often a good strategy to train a model with as much, and as diverse a dataset as possible~\cite{li2012improving}.

In this paper, we consider a highly adverse scenario, where both source and target domain speech are available, but word transcripts are only available for the source domain data.
We present novel augmentation-based methods that transform speech, but, do not require altering existing transcripts. 
Specifically, we first train an unsupervised sequence-to-sequence recurrent variational autoencoder (VAE) on both source and target domain data to learn a latent representation of speech.
We then transform ``nuisance" attributes of speech, such as speaker identities and noise types, that do not contain linguistic information, and are thus irrelevant to ASR, by modifying the latent representation, in order to create additional labeled training data whose distribution is more similar to the target domain.
We evaluate the proposed methods on data from the CHiME-4 challenge~\cite{vincent2016analysis}, which is highly mismatched from the source domain data, WSJ0~\cite{garofalo2007csr}.
The proposed method reduces the absolute word error rate (WER) by as much as 35\% compared to a non-adapted baseline.

The rest of the paper is organized as follows. In Section~\ref{sec:model}, we introduce the VAE model, and present the augmentation methods in Section~\ref{sec:aug}. Related work is discussed in Section~\ref{sec:relate}. Experimental setup and results are shown in Section~\ref{sec:setup} and \ref{sec:res} respectively. Lastly, we conclude our work and discuss about future work in Section~\ref{sec:conclu}.




\section{Variational Autoencoder Model}\label{sec:model}
\subsection{Variational Autoencoder}
Consider a speech dataset $\mathcal{D} = \{\bm{x}^{(i)}\}_{i=1}^N$ consisting of N i.i.d. samples of observed variables $\bm{x}$. We assume that the data are generated from some random process that involves latent variables $\bm{z}$ as follows: 
\begin{enumerate*}[label=(\arabic*)]
	\item a latent variable $\bm{z}$ is drawn from a prior distribution $p_{\theta^*}(\bm{z})$;
    \item an observed variable $\bm{x}$ is drawn from a conditional distribution $p_{\theta^*}(\bm{x}|\bm{z})$.
\end{enumerate*}
We assume that both $p_{\theta^*}(\bm{z})$ and $p_{\theta^*}(\bm{x}|\bm{z})$ come from some distribution family parameterized by $\theta$. To learn this generative process with the presence of only observed data, it is often required to estimate the posterior distribution $p_{\theta}(\bm{z}|\bm{x})$; however, with moderately complex conditional distributions $p_{\theta}(\bm{x}|\bm{z})$, true posterior distributions are generally intractable.

A variational autoencoder~\cite{kingma2013auto} addresses this issue by introducing an encoder and a decoder. The decoder maps the latent variable $\bm{z}$ to the conditional distribution $p_{\theta}(\bm{x}|\bm{z})$, while the encoder maps the observed variable $\bm{x}$ to the approximated posterior distribution $q_{\phi}(\bm{z}|\bm{x})$, an approximation of the true posterior $p_{\theta}(\bm{z}|\bm{x})$. 
The log marginal likelihood of the dataset is the sum of individual log marginal likelihood $\log p_{\theta}(\bm{x}^{(1)}, \bm{x}^{(2)}, \cdots, \bm{x}^{(N)}) = \sum_{i=1}^{N} \log p_{\theta}(\bm{x}^{(i)})$, and each can be rewritten as:
  \begin{align}
      &\log p_{\theta}(\bm{x}) = D_{KL}(q_{\phi}(\bm{z}|\bm{x})||p_{\theta}(\bm{z}|\bm{x})) + \mathcal{L}(\theta,\phi;\bm{x}) \nonumber\\
      &\ge \mathcal{L}(\bm{\theta},\bm{\phi};\bm{x}) \nonumber\\
      &= - D_{KL}(q_{\bm{\phi}}(\bm{z}|\bm{x}) || p_{\bm{\theta}}(\bm{z}) ) +  \mathbb{E}_{q_{\bm{\phi}}(\bm{z}|\bm{x})} [ \log p_{\bm{\theta}}(\bm{x}|\bm{z}) ] \label{eq:lb},
  \end{align}
where $\mathcal{L}(\theta,\phi;\bm{x})$ is the variational lower bound to the log marginal likelihood of data $\bm{x}$.
Since the exact ML estimation for the parameter $\theta$ is intractable, we optimize the variational lower bounds in Eq.\ref{eq:lb} of the dataset instead to obtain the approximate ML estimation for $\theta$. Under certain mild condition for $q_{\phi}(\bm{z}|\bm{x})$, we can rewrite the second term in Eq.\ref{eq:lb} to be the expectation taken over an auxiliary noise distribution such that the Monte Carlo estimation of the expectation is differentiable w.r.t. $\phi$. By maximizing Eq.\ref{eq:lb}, we can apply stochastic gradient based approaches to jointly optimize $\theta$ and $\phi$.

\subsection{Sequence-to-Sequence Recurrent VAE Architecture}
In this work, we use a filter bank for the frame-level representation of speech, which are extracted every 10ms using a 25ms window. We let the observed data $\bm{x} = \{ x_{1}, \cdots, x_{20} \}$ be a sequence of 20 frames, roughly at the scale of a syllable. VAEs are applied to learn the generative process of syllable-level speech segments. For the model we consider here, both the conditional distribution $p_{\theta}(\bm{x}|\bm{z})$ and the approximate posterior distribution $q_{\phi}(\bm{z}|\bm{x})$ are diagonal Gaussian distributions:
\begin{align*}
	p_{\theta}(\bm{x}|\bm{z}) &= \mathcal{N} (\bm{z}; f_{\bm{\mu}_{\bm{z}}}(\bm{x};\theta), \exp(f_{\log\sigma^2_{\bm{z}}}(\bm{x}; \theta))) \\
    q_{\phi}(\bm{z}|\bm{x}) &= \mathcal{N} (\bm{z}; g_{\bm{\mu}_{\bm{x}}}(\bm{z};\phi), \exp(g_{\log\sigma^2_{\bm{x}}}(\bm{z}; \phi))),
\end{align*}
of which the mean ($f_{\bm{\mu}_{\bm{z}}}(\bm{x};\theta)$ and $g_{\bm{\mu}_{\bm{x}}}(\bm{z};\phi)$) and the log variance ($f_{\log\sigma^2_{\bm{z}}}(\bm{x}; \theta)$ and $g_{\log\sigma^2_{\bm{x}}}(\bm{z}; \phi)$) are computed with neural networks. The prior is considered a centered isotropic multivariate Gaussian $p_{\theta}(\bm{z}) = \mathcal{N}(\bm{z}; \bm{0}, \bm{I})$ of 64 dimensions.

\begin{figure}[tbh]
    \centering
      \centerline{\includegraphics[width=1.0\linewidth]{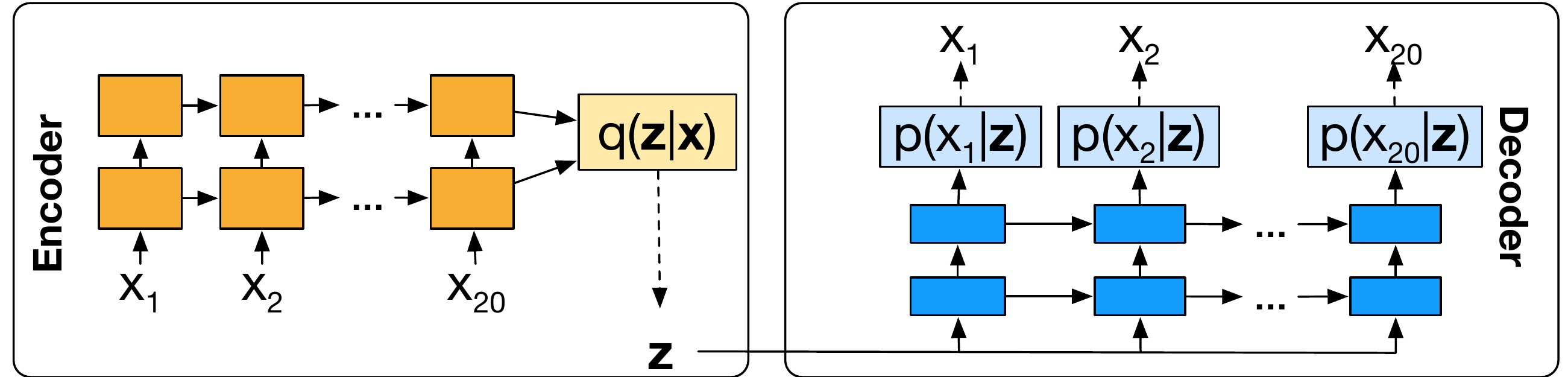}}
    \caption{Illustration of Seq2Seq LSTM VAE architecture.}
    \label{fig:arch}
\end{figure}

To model the temporal relationship within speech segments, we apply a sequence-to-sequence long short-term memory (Seq2Seq-LSTM) architecture as illustrated in Figure~\ref{fig:arch}. 
The encoder is a two layer LSTM with 512 hidden units, which inputs the speech segment frame by frame. 
The outputs from both layers are then concatenated and fed into a fully connected Gaussian parameter layer that predicts the mean and the log variance of the latent variable $\bm{z}$. 
The reparameterization trick is applied to rewrite the latent variable as $\bm{z} = f_{\bm{\mu}_{\bm{z}}}(\bm{x};\theta) + \sqrt{\exp(f_{\log\sigma^2_{\bm{z}}}(\bm{x}; \theta))} \odot \bm{\epsilon}$, where $\odot$ denotes the element-wise product, and vector $\bm{\epsilon}$ is sampled from $\mathcal{N}(\bm{0}, \bm{I})$. 

The decoder is also a two layer LSTM with 512 hidden units that takes the sampled latent variable as the input and generates a sequence of outputs. Each output is used as the input to another fully-connected Gaussian parameter layer that predicts the mean and the log variance for one frame of $\bm{x}$. The entire model can be seen as a stochastic sequence-to-sequence autoencoder that encodes a frame sequence stochastically to the latent space, and then decodes statistically from a sampled latent variable to a sequence of frames.

\subsection{Latent Attribute Representations}
To losslessly reconstruct speech segments from their latent variables, the latent variables must encode the factors that result in the variability of speech segments. 
It is proposed and empirically verified in~\cite{hsu2017learning} that VAEs learn to use orthogonal subspaces to encode speaker and phone attributes, and the prior distribution of $\bm{z}$ conditioned on some label $r$ of some type of attribute $a$ is normally distributed. 
Suppose $y_a$ is the associated label of the attribute $a$ for data $\bm{x}$, these assumptions can be formulated as follows: $p_{\theta}(\bm{z}|y_a = r) = \mathcal{N}(\bm{z}; \bm{\mu}_r, \bm{\Sigma}_r)$, and $\bm{\mu}_{r_i} \perp \bm{\mu}_{r_j}$ if $r_i$ and $r_j$ are labels of different types of attributes, such as a speaker and a phone.

The mean of the conditional prior $\bm{\mu}_r$ is defined as the \textit{latent attribute representation}, and can be estimated by averaging the latent variables of speech segments which have the label $r$. This can be formulated as follows:
\begin{equation}
	\bm{\mu}_r \approx \sum_{i=1}^{N} f_{\bm{\mu}_{\bm{z}}}(\bm{x}^{{(i)}};\theta) \mathbbm{1}_{y_a^{(i)} = r} / \sum_{i=1}^{N} \mathbbm{1}_{y_a^{(i)} = r}. \label{eq:lar}
\end{equation}

\section{Data Augmentation Methods}\label{sec:aug}
Here we introduce the idea of nuisance attribute representations in the scenario of speech recognition, and discuss how to compute these representations without supervision. 
We then summarize how we generate transformed labeled training utterances based on these nuisance attribute representations.

\subsection{Nuisance Attributes and VAE-Based Augmentation Method}
We define the \textit{nuisance attribute} to be the factors that affect the surface form of a speech utterance but not the linguistic content, such as speaker identity, channel, and background noise, etc.
These attributes, unlike phonetic attributes, are generally consistent within an utterance, which implies that we can assume the labels for these attributes are the same for all the segments within an utterance. 
Therefore, we can compute one \textit{latent nuisance representation} for each utterance using Eq.\ref{eq:lar}. 
Suppose $\{\bm{x}_{utt_j}^{(i)}\}_{i=1}^{N_j}$ is the set of segments from an utterance $utt_j$, we then have $\bm{\mu}_{utt_j} = \sum_{i=1}^{N_j} f_{\bm{\mu}_{\bm{z}}}(\bm{x}_{utt_j}^{{(i)}};\theta) / N_j$.

\begin{figure}[tbh]
    \centering
      \centerline{\includegraphics[width=\linewidth]{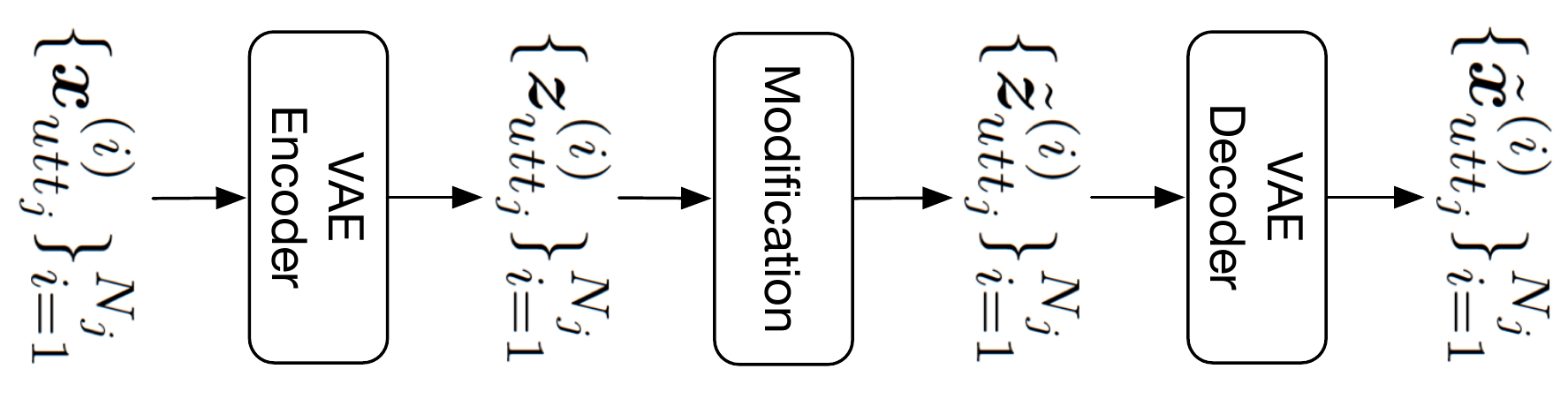}}
    \caption{Flowchart of generating transformed labeled data.}
    \label{fig:flow}
\end{figure}

We generate labeled training data (i.e. with transcripts) for automatic speech recognition systems by transforming nuisance attributes of the labeled source data.
The newly generated data can still use the original transcript for training but differ in some aspect, such as speaker quality and background noise, from the original speech.
Figure~\ref{fig:flow} shows the flowchart of generating transformed labeled data.
Let $(\{\bm{x}_{utt_j}^{(i)}\}_{i=1}^{N_j}, tra_{utt_j})$ be the source utterance of $N_j$ segments with the transcript $tra_{utt_j}$ that we want to modify from.
We first encode and sample each segment $\bm{x}_{utt_j}^{(i)}$ to generate $\bm{z}_{utt_j}^{(i)}$ using a trained VAE encoder.
Then the same modification operation is applied to each latent variable in $\{\bm{z}_{utt_j}^{(i)}\}_{i=1}^{N_j}$ to produce $\{\tilde{\bm{z}}_{utt_j}^{(i)}\}_{i=1}^{N_j}$. 
Finally, we decode $\{\tilde{\bm{z}}_{utt_j}^{(i)}\}_{i=1}^{N_j}$ using the same trained VAE decoder to obtain the modified utterance $\{\tilde{\bm{x}}_{utt_j}^{(i)}\}_{i=1}^{N_j}$ that shares the same transcript $tra_{utt_j}$ with the original utterance.
In other words, we create new labeled training data: $(\{\tilde{\bm{x}}_{utt_j}^{(i)}\}_{i=1}^{N_j}, tra_{utt_j})$.
We next introduce two types of modification operations in Section~\ref{sec:repl} and \ref{sec:pert} respectively.

\subsection{Type I: Nuisance Attribute Replacement}\label{sec:repl}
The first type of modification operation we consider is to replace the nuisance attribute of one utterance with that of another utterance.
We assume VAEs use orthogonal subspaces to model phone attributes and nuisance attributes, and apply the operation derived in~\cite{hsu2017learning}. 
Let $\{\bm{z}_{utt_{src}}^{(i)}\}_{i=1}^{N_j}$ be the encoded latent variables of the source utterance segments we want to modify from, $\bm{\mu}_{utt_{src}}$ be the latent nuisance representation from the source utterance, and $\bm{\mu}_{utt_{tar}}$ be the latent nuisance representation from the target utterance.
Then we modify $\bm{z}_{utt_{src}}^{(i)}$ as follows:
\begin{equation*}
	\tilde{\bm{z}}_{utt_{src}}^{(i)} = \bm{z}_{utt_{src}}^{(i)} - \bm{\mu}_{utt_{src}} + \bm{\mu}_{utt_{tar}}.
\end{equation*}

\begin{figure}[tbh]
	\begin{minipage}[b]{\linewidth}
  		\centering
  		\centerline{\includegraphics[width=\linewidth]{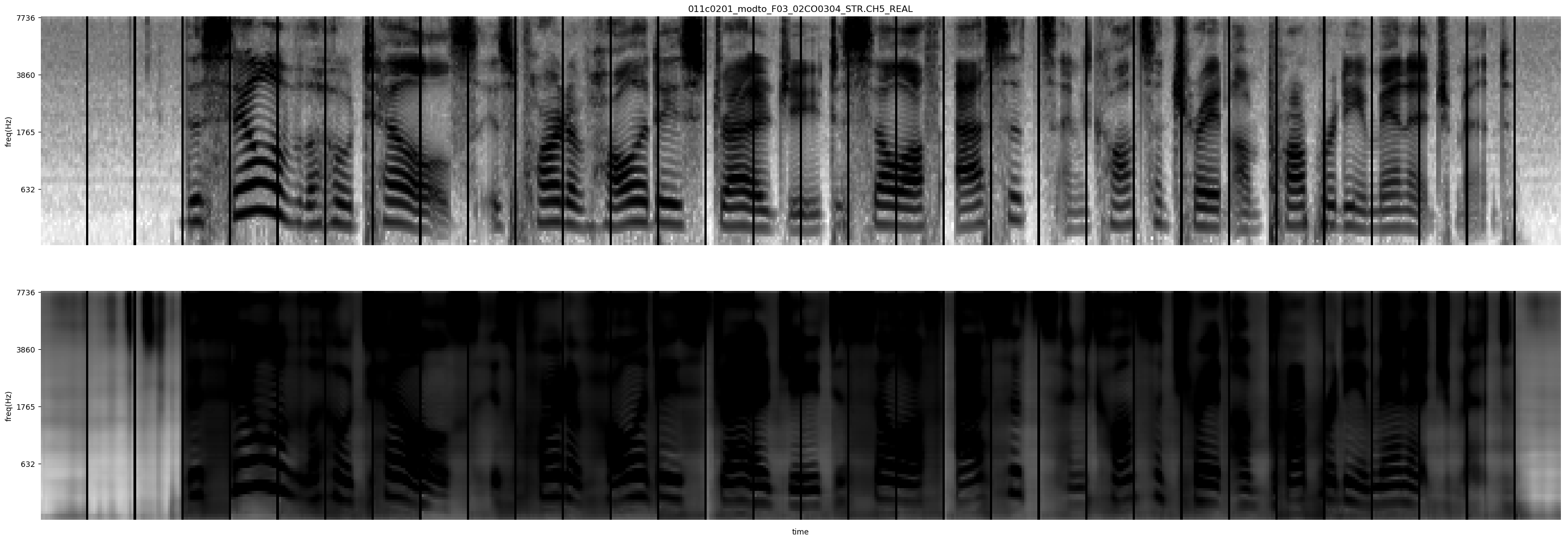}}
  		\centerline{(a) modifying a clean utterance to be like a noisy utterance}
	\end{minipage}
    \begin{minipage}[b]{\linewidth}
  		\centering
  		\centerline{\includegraphics[width=\linewidth]{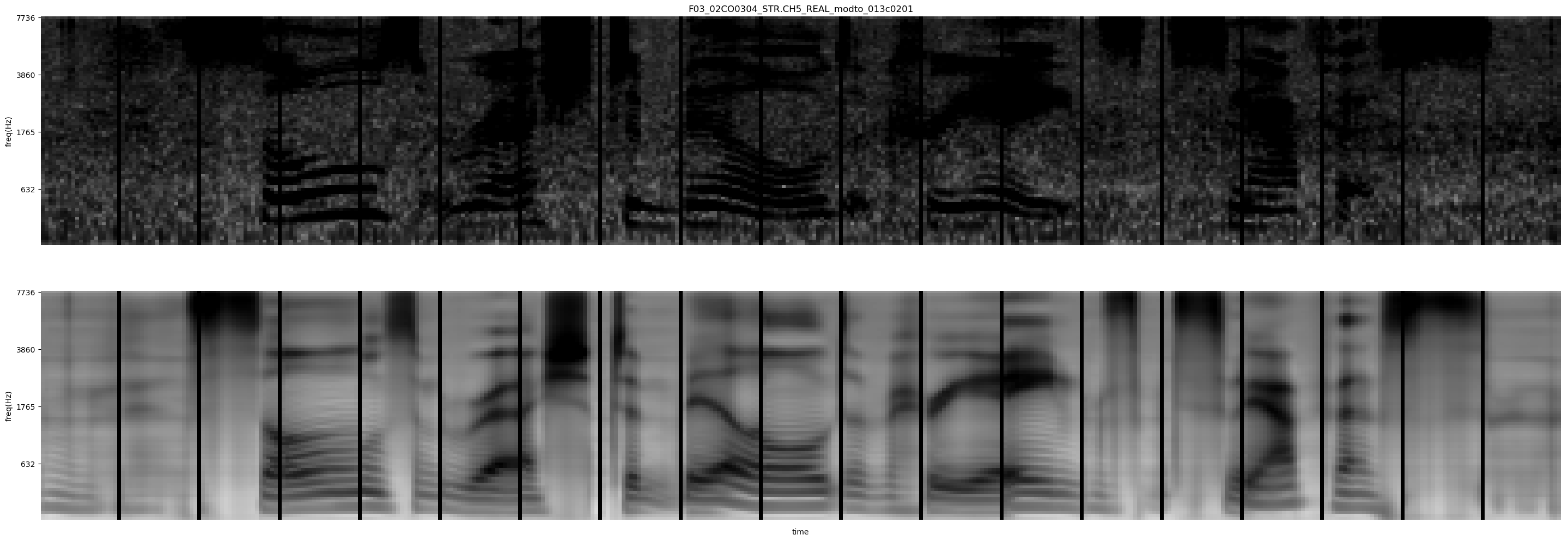}}
  		\centerline{(b) modifying a noisy utterance to be like a clean utterance}
	\end{minipage}
    \caption{Two examples of replacing the nuisance attributes.}\label{fig:repl}
\end{figure}

Figure~\ref{fig:repl} shows two examples of modifying the nuisance attributes, where the first row is the original utterance and the second row is the modified utterance. 
In Figure~\ref{fig:repl}(a), a clean utterance is modified by replacing its nuisance attributes with those from a noisy utterance. 
Conversely, Figure~\ref{fig:repl}(b) illustrates an example of modifying a noisy utterance by replacing its nuisance attributes with those estimated from a clean utterance. 
In the figure, segments within an utterance are separated by vertical black lines. 
From both examples we can observe that while the spacing between harmonics and the level of noise changes, the linguistic content does not seem to change after replacing the nuisance attributes.
 
\subsection{Type II: Latent Nuisance Subspace Perturbation}\label{sec:pert}
The fundamental assumption of this work is that VAEs learn to use orthogonal subspaces to model linguistic factors and nuisance factors respectively. 
Hence, we are able to modify the nuisance attribute without changing the original linguistic attribute by only modifying factors in the latent nuisance subspace, but keeping factors in the latent linguistic subspace intact.
While the operation in Section~\ref{sec:repl} bypasses the search for the latent nuisance subspace, we can alternatively discover this subspace, and then sample or perturb the factors in it to change the nuisance attribute of an utterance.

\subsubsection{Determining the latent nuisance subspace with PCA}
\begin{figure}[tbh]
    \centering
      \centerline{\includegraphics[width=1.0\linewidth]{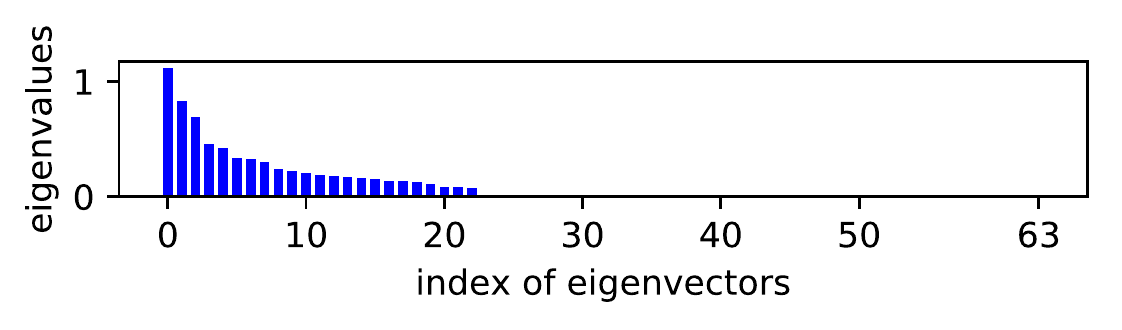}}
    \caption{Eigenvalues of PCA analysis on latent nuisance representations in a descending order.}
    \label{fig:eigval}
\end{figure}

Given a dataset of $M$ utterances, we can compute $M$ latent nuisance representations $\{\bm{\mu}_{utt_j}\}_{j=1}^M$, with one for each utterance.
The latent nuisance subspace is composed of a set of bases, which captures the variations among these latent nuisance representations.
We apply principle component analysis (PCA) on the $M$ latent nuisance representations to obtain a list of eigenvectors $\{\bm{e}_d\}_{d=1}^D$, sorted in a descending order by their associated eigenvalues $\{\sigma^2_d\}_{d=1}^D$, where $D$ is the dimension of the latent variable $\bm{z}$. 
Each eigenvalue interprets the variance of latent nuisance representations along the direction of its associated eigenvector. We plot the eigenvalues in Figure~\ref{fig:eigval} in a descending order, where we can observe that most of the variation is captured by the first few dimensions.

\subsubsection{Soft latent nuisance subspace perturbation}
An intuitive way to determine and perturb the latent nuisance subspace is to select the first few eigenvectors and only perturb in those directions. 
We refer to this as \textit{hard latent nuisance subspace perturbation}, since it demands a hard decision on the rank of the subspace. 
Alternatively, we propose an approach called \textit{soft latent nuisance subspace perturbation}, which generates a \textit{perturbation vector} $\bm{p}$ as follows:
\begin{equation}
	\bm{p} = \gamma \sum_{d=1}^D \psi_d \sigma_d \bm{e}_d, \quad \psi_d \sim \mathcal{N}(0, 1), \nonumber
\end{equation}
where $\psi_d$ is drawn from a normal distribution, $\sigma_d$ and $\bm{e}_d$ are square root of $d$-th largest eigenvalue and its associated eigenvector, and $\gamma$ is a hyper-parameter, referred to as the \textit{perturbation ratio}.
It can be observed that the expected scale we perturb along an eigenvector $\bm{e}_d$ is proportional to the standard deviation of latent nuisance representations along that eigenvector, which is the square root of its eigenvalue $\sigma_d^2$. This approach thus automatically adapts to different distributions of eigenvalues, regardless how many dimensions a VAE learns to use to model the nuisance attributes.

\begin{figure}[tbh]
    \centering
      \centerline{\includegraphics[width=\linewidth]{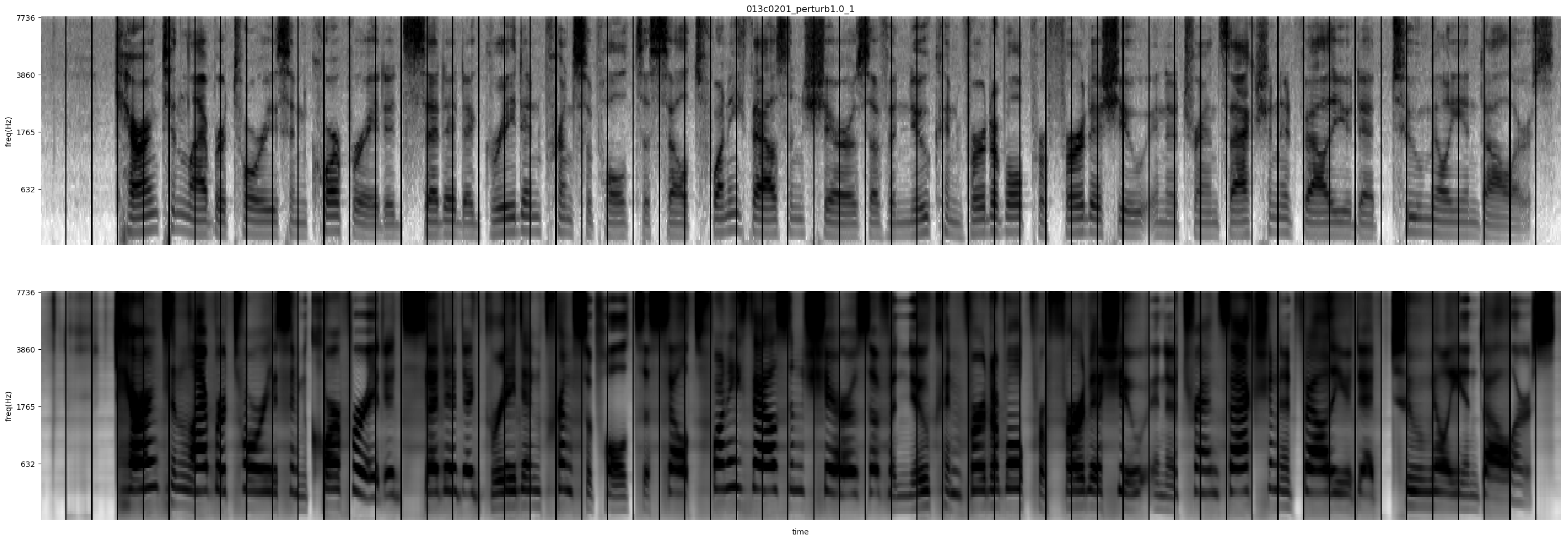}}
    \caption{An example of perturbing latent nuisance attributes.}
    \label{fig:pert}
\end{figure}

Let $\{\bm{z}_{utt_{src}}^{(i)}\}_{i=1}^{N_j}$ be the encoded latent variables of the source utterance segments we want to perturb. We modify each latent variable as follows:
\begin{equation}
	\tilde{\bm{z}}_{utt_{src}}^{(i)} = \bm{z}_{utt_{src}}^{(i)} + \bm{p}, \nonumber
\end{equation}
which adds the same perturbation vector $\bm{p}$ to each segment in an utterance such that the nuisance attribute change is consistent for all segments within an utterance. Figure~\ref{fig:pert} shows an example of perturbing the latent nuisance attributes with $\gamma = 1.0$, where the first row is the original utterance and the second row is the perturbed utterance.

\section{Related Work}\label{sec:relate}
Deep domain adaptation (DDA) to acoustic modeling~\cite{sun2017unsupervised} is a state-of-the-art approach that also addresses the unsupervised domain adaptation problem for robust speech recognition. 
This work adopts a domain-adversarial training method for neural networks~\cite{ganin2016domain}, which encourages the networks to learn features that are discriminative for the main learning task, but not discriminative with respect to domains.
Specifically, the neural network acoustic model in~\cite{sun2017unsupervised} is composed of one common feature extractor network, and two predictor networks that predict the senone labels and domain labels, respectively. 
The feature extractor network and the senone predictor network are trained to minimize the senone prediction error and maximize the domain prediction error, while the domain predictor network is trained to minimize the domain prediction error.
Hence, during training, unlabeled target-domain data are used to update the feature extractor and domain predictor network, and labeled source-domain data are used to update all three networks.

Our proposed data augmentation method takes a different view by generating more domain-diverse data in order to train more robust models.
The two views are complementary, and could be potentially applied in combination.

\section{Setup}\label{sec:setup}


\subsection{Dataset}
Our dataset is based on the CHiME-4 challenge~\cite{vincent2016analysis}, which targets distant-talking ASR and whose setup is based on the speaker-independent medium (5K) vocabulary subset of the Wall Street Journal (WSJ0) corpus~\cite{garofalo2007csr}.
The training set of the CHiME-4 dataset consists of 1,600 utterances recorded in four noisy environments from four speakers, and 7,138 simulated noisy utterances based on the clean utterances in the WSJ0 SI-84 training set.
We use the original 7,138 clean utterances as the labeled source-domain data, and the 1600 single channel real noisy utterances as the unlabeled target-domain data for unsupervised domain adaptation.
Performance is evaluated on both the real noisy utterances and the original clean utterances in the development partition of the CHiME-4 dataset in terms of the word error rate (WER).

In addition, we also repeat our experiments on Aurora-4~\cite{pearce2002aurora} to compare with the results reported in~\cite{sun2017unsupervised}. Aurora-4 is a broadband corpus designed for noisy speech recognition tasks based on WSJ0 as well. Two microphone types, clean/channel are included, and six noise types are artificially added to both microphone types, which results in four conditions: clean(A), channel(B), noisy(C), and channel+noisy(D). We use the clean training set as the labeled source-domain data, and the multi-condition development set as the unlabeled target-domain data. The multi-condition test\_eval92 set is used for evaluation.

\subsection{VAE Setup and Training}
All the original clean utterances and the real noisy utterances are mixed and split into training and development sets with the ratio of 90-10 for training the Seq2Seq LSTM VAE.
The VAE is trained with stochastic gradient descent using a mini-batch size of 128 without clipping to minimize the negative variational lower bound plus an $L2$-regularization with weight $10^{-4}$. The Adam~\cite{kingma2014adam} optimizer is used with $\beta_1=0.95$, $\beta_2=0.999$, $\epsilon=10^{-8}$, and initial learning rate of $10^{-3}$. Training is terminated if the lower bound on the development set does not improve for 50 epochs.

\subsection{ASR Setup and Training}
Kaldi~\cite{povey2011kaldi} is used for feature extraction, decoding, forced alignment, and training of an initial HMM-GMM model on the original clean utterances. 
The recipe provided by the CHiME-4 challenge (\texttt{run\_gmm.sh}) and the Kaldi Aurora-4 recipe are adapted by only changing the training data being used.
The Computational Network Toolkit (CNTK)~\cite{yu2014introduction} is used for neural network-based acoustic model training.
For all CHiME-4 experiments, the same LSTM acoustic model~\cite{sak2014long} with the architecture proposed in~\cite{zhang2016highway} is applied, which has 1,024 memory cells and a 512-node projection layer for each LSTM layer, and 3 LSTM layers in total.
Following the training setup in~\cite{hsu2016exploiting}, LSTM acoustic models are trained with a cross-entropy criterion, using truncated backpropagation-through-time (BPTT)~\cite{williams1990efficient} to optimize. 
Each BPTT segment contains 20 frames, and each mini-batch contains 80 utterances, since we find empirically paralleling 80 utterances has similar performance to 40 utterances. 

For all Aurora-4 experiments, the same 6 layer fully-connected deep neural network (DNN) acoustic model with 2,048 hidden units at each layer is applied, which is same architecture as the one in~\cite{sun2017unsupervised}, except that the number of hidden units are doubled. 
The input to the DNN is a context window of 11 frames, with five frames of left context and five frames of right context. 
Each frame is represented using filter bank features with delta and delta-delta coefficients as proposed in~\cite{sun2017unsupervised}.
DNN acoustic models are trained with the cross-entropy criterion, with a mini-batch size of 256.
For both LSTM and DNN training, a momentum of 0.9 is used starting from the second epoch~\cite{zhang2015speech}. 
Ten percent of the training data is held out as a validation set to control the learning rate. The learning rate is halved when no gain is observed after an epoch.

We assume for both nuisance attribute replacement and latent nuisance subspace perturbation, the time alignment of senones does not change.
Therefore, the same forced alignment is used to train the acoustic models.

\section{Experimental Results and Discussion}\label{sec:res}
\begin{table*}[tbh]
  \centering
  \begin{tabular}{|lll|cc|cccc|}
    \hline
    \multicolumn{3}{|c}{Setting}	& \multicolumn{2}{|c|}{WER (\%)} & \multicolumn{4}{c|}{WER (\%) in Noisy Condition by Type} \\
    Exp. Index	& Aug. Method & Fold 		& Clean		& Noisy		& BUS	& CAF	& PED	& STR	\\
    \hline
	\multirow{2}{*}{1}	& Orig.		& 1		& \textbf{19.04}  	& \textbf{87.80}	& 96.16	& 92.35	& 78.46	& 84.24	\\
    					& Recon.		& 1		& 19.61  	& 90.72	& 98.95	& 93.45	& 81.52	& 88.97	\\
    \hline\hline
    \multirow{2}{*}{2}	& Repl. Clean	& 1 	& \textbf{20.03} 	& 67.12  & 71.99	& 76.84	& 55.32 	& 64.33	\\
    					& Repl. Noisy	& 1    	& 26.31 	& \textbf{57.66} 	& 62.12   & 69.25 	& 46.89  & 52.38 	\\
    \hline\hline
	\multirow{3}{*}{3}	& Pert., $\gamma=1.0$    		& 1		& 20.01	& \textbf{53.06}	& 55.66	& 66.12	& 41.94	& 48.50 	\\
    					& Uni-Pert., $\gamma=1.0$   	& 1		& \textbf{19.70}	& 65.07	& 69.27	& 75.28	& 53.65 	& 62.06	\\
    					& Rev-Pert., $\gamma=1.0$   	& 1		& 19.75	& 87.98	& 95.13	& 90.58	& 76.71	& 89.50 	\\
    \hline\hline
    \multirow{4}{*}{4}	& Pert., $\gamma=0.5$    	& 1		& \textbf{19.55}	& 65.61	& 67.87	& 77.37	& 54.54 	& 62.66	\\
    					& Pert., $\gamma=1.0$    	& 1		& 20.01	& \textbf{53.06}	& 55.66	& 66.12	& 41.94	& 48.50 	\\
    					& Pert., $\gamma=1.5$    	& 1		& 19.99	& 53.59	& 57.09	& 64.91	& 42.23 	& 50.11	\\
    					& Pert., $\gamma=2.0$    	& 1		& 20.39	& 58.10	& 64.35	& 69.12	& 45.39 	& 53.55	\\
    \hline\hline
	\multirow{3}{*}{5}	& Orig. $+$ Repl. Noisy	& 2		& 19.88	& 55.72	& 60.72	& 66.46	& 45.08 	& 50.63	\\
    					& Repl. Noisy				& 2		& 25.26	& 55.59	& 59.24	& 67.85	& 44.65 	& 50.63	\\
    					& Pert., $\gamma=1.0$		& 2		& \textbf{19.82}	& \textbf{52.49}	& 55.52	& 65.04	& 41.17 	& 48.24	\\
    \hline
  \end{tabular}
  \caption{CHiME-4 development set word error rate of acoustic models trained on different augmented sets.}
  \label{tab:wer}
\end{table*}

In this section, we verify the effectiveness of the proposed VAE-based data augmentation methods for unsupervised domain adaptation.
On each dataset, the same acoustic model architectures and training procedures, as well as the same language models are used for all the experiments. 
For the CHiME-4 dataset, besides reporting the WER on the clean and the noisy development sets respectively, we also show the WER for the noisy set by the four recording locations: bus (BUS), cafe (CAF), pedestrian area (PED), and street junction (STR).  All the CHiME-4 results are listed in Table~\ref{tab:wer}. 
For the Aurora-4 dataset, we report the averaged WER as well as the WER in four conditions in Table~\ref{tab:a4_wer}.
Different sets of experiments are separated by double horizontal lines and indexed by the \textit{Exp. Index} on the first column.
The second column, \textit{Aug. Method}, explains the augmentation method and the hyper-parameter being used. The ratio of the new training set to the original clean training set is listed on the third column, referred to as the \textit{Fold}.

\subsection{Baselines}
We first establish baselines by training models on two sets.
The first set, \textit{Orig.}, refers to the original clean training set that does not involve VAE.
The second set, \textit{Recon.}, refers to the reconstructed clean training set that is generated by using the VAE to first encode and then decode.
Note that this does not involve the modification methods mentioned in Sections~\ref{sec:repl} and \ref{sec:pert}.

The results are listed in Table~\ref{tab:wer}, \textit{Exp. Index 1}. 
The fourth row shows the results on the matched domain (clean), and the fifth row shows the results on the mismatched domain (noisy). 
It can be observed that the performance degrades significantly when the models are tested on the mismatched domain.
The WER increases from 19.04\% to 87.08\% for \textit{Orig.}, and from 19.61\% to 90.72\% for \textit{Recon.} respectively.
In addition, since the reconstruction from the VAE is not perfect, part of the information may be lost during this process.
Hence, the model trained on \textit{Recon.} is slightly worse than the one trained on \textit{Orig.} for all testing conditions.
Lastly, the relative WERs of the four location are consistent on the both training sets. BUS appears to be the most difficult one, while PED is the easiest one among the four locations.


\subsection{Replacing Nuisance Attributes}
We evaluate the effectiveness of augmenting data by replacing the nuisance attributes as mentioned in Section~\ref{sec:repl}.
Let $\mathcal{U}_{src} = \{ \bm{\mu}_{utt_j} \}_{j=1}^{M_{src}}$ be the set of latent nuisance representations of the source domain utterances, and $\mathcal{U}_{tar} = \{ \bm{\mu}_{utt_j} \}_{j=M_{src}}^{M}$ be the set of latent nuisance representations of the target domain utterances. 
$M_{src}$ is the number of source domain utterances, and $M_{tar} = M - M_{src}$ is the number of target domain utterances.
We create the augmented set \textit{Repl. Clean} by replacing the latent nuisance representation of each source domain utterance with one drawn from $\mathcal{U}_{src}$.
The \textit{Repl. Noisy} is generated similarly but is replaced with one drawn from $\mathcal{U}_{tar}$. 

The results are shown in Table~\ref{tab:wer}, \textit{Exp. Index 2}.
For both augmented methods, we observe at least 20\% absolute WER reduction on the target domain compared to the baselines. 
We observe an additional 10\% absolute WER reduction when replacing the latent nuisance representations with those taken from the target domain instead of the source domain.  We also observe that \textit{Repl. Noisy} shows 6\% worse WER on the source domain than \textit{Repl. Clean}.
The relative strength of \textit{Repl. Clean} and \textit{Repl. Noisy} on different domains verifies the effectiveness of our proposed method at shifting the distribution from one domain to another. 


\begin{table*}[tbh]
  \centering
  \begin{tabular}{|lll|l|llll|}
    \hline
    \multicolumn{3}{|c}{Setting}	& \multicolumn{1}{|c|}{WER (\%)} & \multicolumn{4}{c|}{WER (\%) by Condition} \\
    Exp. Index	& Aug. Method/Baselines & Fold 		& Avg.	& Clean(A)	& Noisy(B)	& Channel(C)	& Channel+Noisy(D)	\\
    \hline
    \multirow{2}{*}{0}	& Clean-DNN-HMM~\cite{sun2017unsupervised}		& -		& 36.22	& 3.36	& 29.74	& 21.02	& 50.73	\\
    					& DDA-DNN-HMM~\cite{sun2017unsupervised}		& -		& 22.53 & 3.24	& 14.52	& 17.82	& 34.55	\\
                        & DNN-PP~\cite{du2014robust}					& - 	& 18.7	& 5.1	& 12.0	& 10.5	& 29.0	\\
    \hline\hline
	\multirow{2}{*}{1}	& Orig.		& 1		& 53.98	& 3.38	& 50.56	& 42.67	& 67.70	\\
    					& Recon.	& 1		& 66.29 	& 4.58	& 65.44	& 51.02	& 79.97	\\
    \hline\hline
    \multirow{1}{*}{2}	& Repl. Noisy				& 1 	& 22.53 & 4.80	& 16.31	& 14.72	& 32.99	\\
    \hline\hline
    \multirow{8}{*}{3}	& Pert., $\gamma=0.5$    	& 1		& 35.37 & 4.11	& 27.73	& 33.51	& 48.52	\\
    					& Pert., $\gamma=1.0$    	& 1		& 24.82 & 4.35	& 17.11	& 22.38	& 36.36	\\
    					& Pert., $\gamma=1.5$    	& 1		& 21.98 & 4.24	& 15.08	& 16.87	& 32.69	\\
    					& Pert., $\gamma=2.0$    	& 1		& 20.68 & 4.45	& 14.33	& 14.74	& 30.72	\\
                        & Pert., $\gamma=2.5$    	& 1		& 20.99 & 4.99	& 15.35	& 15.54	& 30.22	\\
                        & Pert., $\gamma=3.0$    	& 1		& 21.18 & 5.29	& 15.47	& 15.71	& 30.45	\\
                        & Pert., $\gamma=3.5$    	& 1		& 21.33	& 5.45	& 16.13	& 14.70	& 30.29 \\
                        & Pert., $\gamma=4.0$    	& 1		& 22.00 & 6.43	& 17.15	& 15.00	& 30.62	\\
    \hline\hline
	\multirow{4}{*}{4}	& Pert., $\gamma=2.0$		& 2		& 20.06 	& 4.13	& 13.85	& 14.96	& 29.77	\\
                        & Pert., $\gamma=2.0$		& 4		& 19.42	& 4.09	& 13.34	& 14.14	& 28.92 \\
                        & Pert., $\gamma=2.0$		& 8		& 18.86	& 4.28	& 12.89	& 13.51	& 28.16 \\
                        & Pert., $\gamma=2.0$		& 16	& \textbf{18.76}	& 4.04	& 12.84	& 13.54	& 28.01 \\

    \hline
  \end{tabular}
  \caption{Aurora-4 test\_eval92 set word error rate of acoustic models trained on different augmented sets.}
  \label{tab:a4_wer}
\end{table*}

\subsection{Correctness of Soft Latent Nuisance Subspace Perturbation}
We first examine the correctness of our proposed soft latent nuisance subspace perturbation by proposing two alternative perturbation methods.
To eliminate the effect of the perturbation scale on the performance,
we consider two alternative methods subject to the constraint that the expected squared Euclidean norm of the perturbation vector $\bm{p}$ is the same as the proposed method.

Recall that $\bm{p} = \gamma \sum_{d=1}^{D} \psi_d \sigma_d \bm{e}_d$ for our proposed method. We then have: $\mathop{\mathbb{E}} \left[ ||\bm{p}||_2^2 \right] = \gamma^2 \sum_{d=1}^{D} \sigma_d^2 \mathop{\mathbb{E}} \left[ \psi_d^2 \right] =  \gamma^2 \sum_{d=1}^{D} \sigma_d^2$.
We then consider \textit{uniform perturbation} to be a method that perturbs each direction with the same expected scale, controlled by the same perturbation ratio hyper-parameter $\gamma$.
The perturbation vector $\bm{p}_{uni}$ derived from this method can be formulated as $\bm{p}_{uni} = \gamma \sum_{d=1}^{D} \psi_d \sigma_{uni} \bm{e}_d$, where $\sigma_{uni} = \sqrt{\sum_{d=1}^{D} \sigma_d^2 / D}$.
In addition, we design another method that reverses the expected perturbation scales from the proposed soft latent nuisance subspace perturbation method, named the \textit{reverse soft latent nuisance subspace perturbing}. 
The perturbation vector $\bm{p}_{rev}$ can be formulated as $\bm{p}_{rev} = \gamma \sum_{d=1}^{D} \psi_d \sigma_{D-d} \bm{e}_d$.

We show the results of soft latent nuisance subspace perturbation (Pert.), uniform perturbation (Uni-Pert.), and reverse soft latent nuisance subspace perturbation (Rev-Pert.) in Table~\ref{tab:wer}, \textit{Exp. Index 3}, which all use the same perturbation ratio $\gamma = 1.0$. 
We can clearly observe the superiority of the proposed method among the three methods applied on the target domain.
Pert.~reduces the absolute WER by almost 35\% from the baseline and outperforms Uni-Pert. by 12\%, while Rev-Pert. achieves almost no improvement.
This experiment verifies the importance of determining an appropriate way to perturb the latent space and the correctness of our method.

\subsection{Effect of Perturbation Ratios}
We next examine the effect on the hyper-parameter $\gamma$ by choosing four scales: 0.5, 1.0, 1.5, 2.0, and list the results in Table~\ref{tab:wer}, \textit{Exp. Index 4}. 
We observe different WER trends for different perturbation ratios for the two testing conditions.
First, regarding the target domain WER, we notice that $\gamma = 1.0$ reaches the best performance among the four scales.
The smaller the perturbation ratio is, the more similar to the original clean data the augmented perturbed data would be.
Hence, when we decrease the perturbation ratio, the performance would asymptotically approach those of the original clean data.
On the other hand, as we increase the perturbation ratio, the chance of the perturbed utterances becoming linguistically different increases.
This may hurt the performance and cancel out the benefit of having more diverse data by perturbing the nuisance attributes.
As for the source domain WER, we observe degradation when increasing the perturbation ratio, because the perturbed data distribution becomes less similar to original clean data distribution.

\subsection{Effect of Dataset Size}
In this section, we study the effect of the size by combining different sets of augmented data or the original data.
Specifically, three cases are considered:
\begin{enumerate*}[label=(\arabic*)]
	\item combining \textit{Repl. Noisy} with the original data.
    \item combining \textit{Repl. Noisy} with another copy of \textit{Repl. Noisy}.
    \item combining \textit{Pert., $\gamma = 1.0$} with another copy of \textit{Pert., $\gamma = 1.0$}.
\end{enumerate*}

The results are listed in Table~\ref{tab:wer}, \textit{Exp. Index 5}.
In the first two cases, both source and target domain WERs are improved from the one-fold \textit{Repl. Noisy}.
While adding another copy of \textit{Repl. Noisy} shows slightly better (0.13\%) WER in the target domain of the first two cases, adding the original clean data significantly reduces (6.43\%) WER in the source domain.
This suggests that the second case addresses the issue of \textit{Repl. Noisy} on shifting the data distribution entirely to the target domain.
In the third case, a slight but consistent 0.19\% and 0.57\% WER reductions from the one-fold \textit{Pert., $\gamma = 1.0$} in the source and target domain are observed.
In summary, all three cases show improvement by increasing the size.

\subsection{Comparing with DDA on Aurora-4}
In this section, we repeat the experiments on the Aurora-4 dataset and compare with deep domain adaptation.
Table~\ref{tab:a4_wer} listed our Aurora-4 results and the reference results.
DNN-PP~ \cite{du2014robust} is a method compared in~\cite{sun2017unsupervised} that requires parallel clean-noisy data for training an speech enhancement model as a preprocessor.

Baseline results are established in Table~\ref{tab:a4_wer}, \textit{Exp. Index 1}, where the models are trained with the original clean features (\textit{Orig.}), and the VAE-reconstructed clean features (\textit{Recon.}), respectively. 
Here we can observe significant degradation on mismatched domains (B, C, and D) from the matched domain (A). 
Results of nuisance attribute replacement and soft latent nuisance subspace perturbation with different perturbation ratios are shown in Table~\ref{tab:a4_wer}, \textit{Exp. Index 2} and \textit{Exp. Index 3}.
Both augmentation methods achieve roughly 30\% absolute WER reduction, and the soft latent nuisance subspace perturbation reaches the best performance when using a perturbation ratio $\gamma=2.0$.
By increasing the dataset size, we observe further WER reduction from two-fold to 16-fold.

Since the detailed training recipe is not provided in~\cite{sun2017unsupervised}, we could not reproduce exactly the same baseline results. However, despite the fact that our baseline is 17.76\% worse than that in~\cite{sun2017unsupervised}, our best system (18.76\%) still achieves better performance than the result of DDA (22.53\%) that was reported in~\cite{sun2017unsupervised}, and matches the results of DNN-PP~\cite{du2014robust} without the need for parallel data.

\section{Conclusion and Future Work}\label{sec:conclu}
In this paper, we present two VAE-based data augmentation methods for unsupervised domain adaptation to robust ASR.
In particular, we study the latent representations obtained from VAEs, which enable us to transform nuisance attributes of speech through modifying the latent variables.
Our proposed methods are evaluated two datasets, and achieve about 35\% absolute WER reduction on both sets.
For future work, we plan to investigate the proposed augmentation method using more advanced FHVAE models~\cite{hsu2017unsupervised}, which explicitly disentangle sequence and segment level attributes in the latent space.

\vfill
\pagebreak


\newpage
\newpage

\bibliographystyle{IEEEbib}
\bibliography{refs.bib}

\end{document}